\pdfoutput=1

\documentclass[11pt]{article}

\usepackage{ACL2023}

\usepackage{times}
\usepackage{latexsym}

\usepackage[T1]{fontenc}

\usepackage[utf8]{inputenc}

\usepackage{microtype}

\usepackage{inconsolata}

\usepackage{graphicx}
\usepackage{amsfonts}
\usepackage{amsmath}
\usepackage{multirow}
\usepackage{booktabs}
\usepackage{makecell}
\usepackage{arydshln}
\usepackage{xcolor}
\usepackage{subfigure}
\usepackage{caption}
\usepackage{ulem}
\usepackage{soul}
\newcommand{\red}[1]{\textcolor{red}{#1}}

\newcommand{\tba}[1]{\cellcolor{red!50}{#1}}
\newcommand{\tbb}[1]{\cellcolor{red!40}{#1}}
\newcommand{\tbc}[1]{\cellcolor{red!30}{#1}}
\newcommand{\tbd}[1]{\cellcolor{red!20}{#1}}
\newcommand{\tbe}[1]{\cellcolor{red!10}{#1}}

\newcommand{\casefont}{\fontsize{8pt}{\baselineskip}\selectfont}

\title{
A Frustratingly Easy Improvement for Position Embeddings \\via Random Padding} 

\author{Mingxu Tao$^{\text{1,2}}$, Yansong Feng$^{\text{1,3}}$, Dongyan Zhao$^{\text{1,2}}$ \\
        $^{\text{1}}$Wangxuan Institute of Computer Technology, Peking University, China \\
        $^{\text{2}}$Center for Data Science, Peking University, China \\
        $^{\text{3}}$The MOE Key Laboratory of Computational Linguistics, Peking University, China \\
        \texttt{\{thomastao, fengyansong, zhaody\}@pku.edu.cn} \\}

\begin{document}
\maketitle
\begin{abstract}

Position embeddings, encoding the positional relationships among tokens in text sequences, make great contributions to modeling local context features in Transformer-based pre-trained language models. However, in Extractive Question Answering, position embeddings trained with instances of varied context lengths may not perform well as we expect. Since the embeddings of rear positions are updated fewer times than the front position embeddings, the rear ones may not be properly trained. In this paper, we propose a simple but effective strategy, \textit{Random Padding},
without any modifications to architectures of existing pre-trained language models. We adjust the token order of input sequences when fine-tuning, to balance the number of updating times of every position embedding. Experiments show that \textit{Random Padding} can significantly improve model performance on the instances whose answers are located at rear positions, especially when models are trained on short contexts but evaluated on long contexts.  Our code and data will be released for future research.

\end{abstract}

\section{Introduction}






Pre-trained language models~\citep{devlin-etal-2019-bert,liu-etal-2019-roberta,Clark2020ELECTRA,joshi-et-al-2020-spanbert} have achieved great success in various natural language processing tasks, including text classification, relation extraction, and 
extractive question answering~(QA). These models with Transformer architectures~\citep{transformer} have powerful ability to model local context, which plays a vital role in question answering~\citep{salant-berant-2018-contextualized,peters-etal-2018-deep}. Compared to models of text classification whose predictions are mainly based on sentence-level representations, many extractive QA models have to determine the start and end boundaries of answer spans from all tokens in context. 
Extractive QA models tend to pick out answer from the neighbour words of overlapping words between context and questions~\citep{jia-liang-2017-adversarial, sugawara-etal-2018-makes}, thus should be sensitive to the relative positions of words in context, 
Therefore, 
how to represent the positional relationships among words is important to a QA model.

Transformer-based models merely employ position embeddings to identify the order of tokens, thus encode the positional relationships among tokens. 
Many popular Transformer-based models, like BERT~\citep{devlin-etal-2019-bert} employs absolute position embedding, which can be considered as a set of learnable vectors.
For each token, its corresponding position embedding will be appended to its token embedding to form the final input representation. Therefore, when fine-tuning, if the token sequence is shorter than the maximum input length (e.g., 512 for BERT), the rear part of position embedding vectors will not be equally updated as the front ones. In practice, it is impossible that all instances have a length exactly equal to the maximum input length of pre-trained language models, thus, the rear part of position embeddings might be updated fewer times than the front ones. 
This may prevent QA models from producing more accurate representations for tokens of answers that are located at the end of the context. 


Many recent studies also concentrate on the topic of \textit{Train Short, Test Long}~\citep{su-etal-2021-RoFormer,press2022train,sun2022length}. They adopt relative position embeddings to prevent the situation that part of absolute position embeddings cannot be updated. However, these methods need to modify the attention mechanism of Transformer layers and pre-train the language models again. Unlike these studies, we focus on to enhance existing models with absolute position embeddings. 


In this paper, we first conduct a pilot experiment in extractive QA to show how insufficient training will affect the quality of position embeddings. 
We propose a simple but effective method, \textit{Random Padding}. We reduce the updating times of the front part of position embeddings, and make the rear ones to be updated more times, via randomly moving padding tokens. To examine whether \textit{Random Padding} can be effective to improve QA models, we first experiment in  a severe scenario, where models are trained on instances with shorter context but have to perform inference over longer contexts. We also experiment on datasets whose training and test sets have similar distributions of context lengths. Experimental results show \textit{Random Padding} can improve QA models in both scenarios. We further provide empirical analyses, which reveal that \textit{Random Padding} plays an important role to improve the representations of rear tokens in an input sequence, via making them better to model positional features of local contexts. 

Our main contributions are as follows:
(1) We propose a simple but effective method, named \textit{Random Padding} to improve 
 the embeddings of rear positions, especially when the model is trained on short contexts. 
%
(2) We inject \textit{Random Padding} into popular Transformer-based QA models without any modification to their architectures, and help the models to predict more accurately when answers are located at rear positions.  
(3) We further reveal that \textit{Random Padding} can improve models on public benchmark datasets of extractive QA and document-level relation extraction.







\section{Background}
\label{sec2_pilot}





\subsection{Task Definition}

In extractive question answering, a model should extract a text span from a given 
passage or document to answer the question. The given
passage or document is also called as \texttt{context}. Formally, 
the extractive QA task can be defined as:
Given a \texttt{question} $Q$ and a \texttt{context} $C=\left<c_0,\,\cdots,\,c_{N_c-1}\right>$, the model should select a text span $\left<c_s,\,\cdots,\,c_e\right>$~($0\leq s\leq e<N_c$) from $C$ to answer $Q$, where $N_c$ is the number of words in $C$, $s$ and $e$ are the start and end boundaries of the answer, respectively.

\subsection{Investigated Model}
\label{sec:inv_model}

Pre-trained language models~(PLMs) 
usually receive a sequence of tokens as input.
Here, we take BERT~\citep{devlin-etal-2019-bert} as an example to introduce how to compose the input of PLM. Following BERT's original pre-processing, we utilize special token \texttt{[CLS]} and \texttt{[SEP]} to separate question and context. Since BERT has to receive input sequences with a fixed length, the input sequences shorter than maximum length will be appended by padding tokens~(\texttt{[PAD]}). Therefore, in vanilla BERT, the input sequence containing the given question and context can be composed as:
$$\makebox{\texttt{[CLS]}~\textit{question}~\texttt{[SEP]}~\textit{context}~\texttt{[SEP]}~\textit{paddings}}$$

\begin{figure}[t]
\begin{center}
\centerline{\includegraphics[width=\columnwidth]{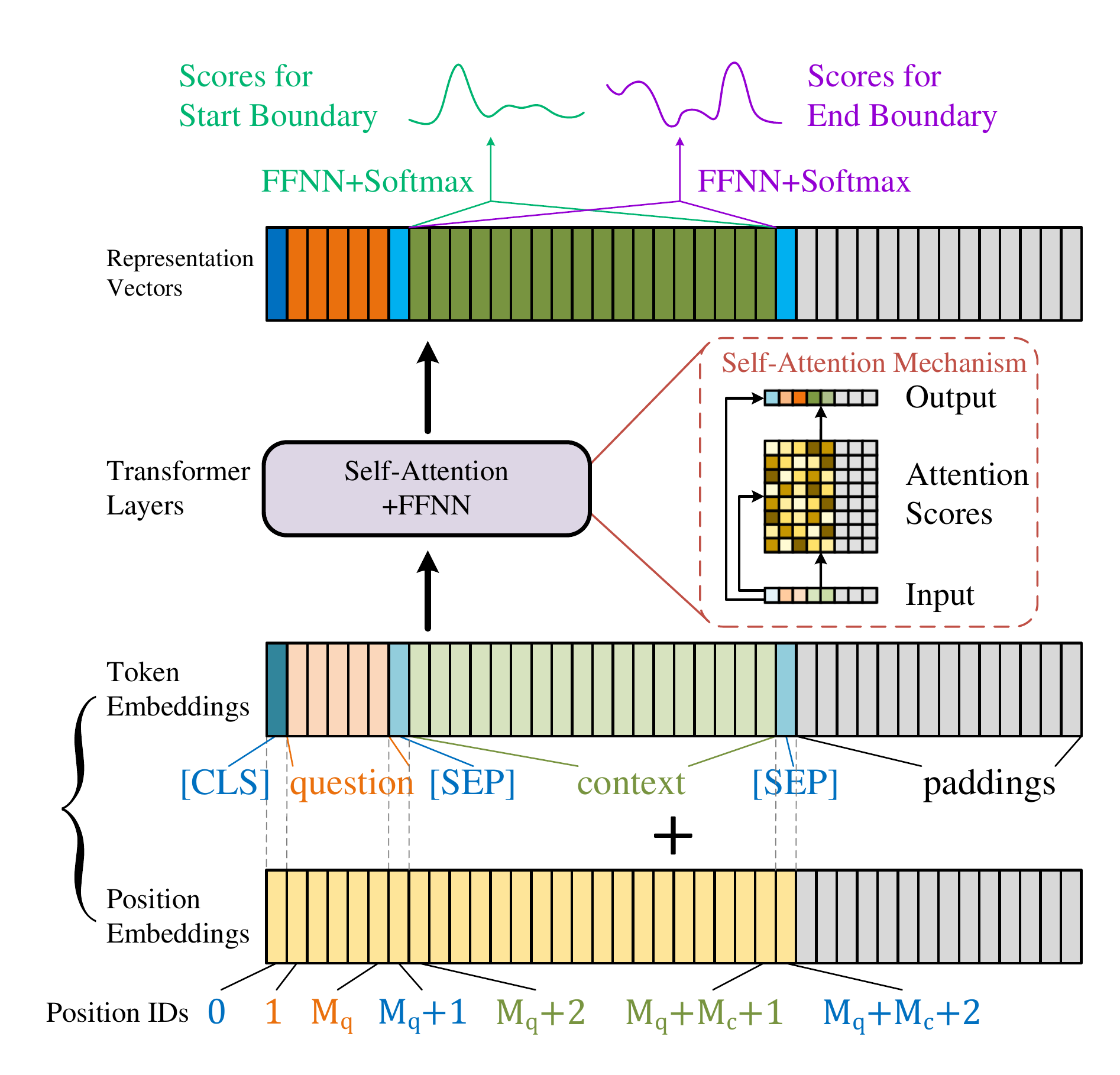}}
\caption{The process of QA model with absolute position embeddings. The grey parts represent the weights corresponding to masked tokens, whose gradients cannot be back-propagated. The representation vectors, embedding vectors and attention scores of non-padding tokens are shown by coloured areas.}  
\label{fig:QA_model} 
\end{center}
\vskip -0.3in
\end{figure}

Formally, assume the question and the context consist of $M_q$ and $M_c$ tokens after tokenization, respectively. Then, we can regard \texttt{[CLS]} located at the $0$-th position, and correspondingly the context includes tokens located at from $M_q+2$ to $M_q+M_c+1$. For ease of exposition, we denote the number of non-padding tokens as $m=M_q+M_c+3$ and the maximum input length as $n$. Thus, there should be $(n-m)$ padding tokens.

In a general extractive QA framework using PLM~\citep{devlin-etal-2019-bert,liu-etal-2019-roberta,Clark2020ELECTRA}, for an input sequence with $m$ non-padding tokens, we denote their representation vectors as $\left\{T_i\right\}_{i=0}^{m-1},\ T_i\in\mathbb{R}^H$. We then employ two trainable vectors $S,E\in\mathbb{R}^H$ to predict the start and the end boundaries of an answer span. The probability of token $s$ being the start boundary of  an answer span can be written as:
$${\rm Prob}({\rm start}=s)=\frac{\exp(S\cdot T_s)}{\sum_{i=M_q+2}^{M_q+M_c+1}\exp (S\cdot T_i)}.$$
Similarly, the probability of token $j$ being the end boundary will be:
$${\rm Prob}({\rm end}=e)=\frac{\exp(E\cdot T_e)}{\sum_{i=M_q+2}^{M_q+M_c+1}\exp (E\cdot T_i)}.$$

Finally, the score of an answer span from the $s$-th token to the $e$-th token can be simplified as $S\cdot T_s+E\cdot T_e$. The ultimate goal of an extractive QA model is to select the maximal $S\cdot T_s+E\cdot T_e$ from any $\left<s,e\right>$ pair with a constraint of $M_q+2\leq s\leq e<M_q+M_c+2$.

Figure \ref{fig:QA_model} illustrates the process of an extractive QA framework based on PLM with absolute position embeddings. As previously mentioned, we add padding tokens behind short input sequences and mask these tokens during fine-tuning. The gradients corresponding to masked tokens will not be back-propagated~(the grey parts in Figure \ref{fig:QA_model}).

\subsection{Pilot Experiment}
\label{sec:pilot_exp}


We observe that the instances in a QA dataset have various question and context lengths, for example, SQuAD~\citep{rajpurkar-etal-2016-squad}. The combined token sequence of question and context after pre-processing are usually shorter than the maximum input length of popular PLM like BERT. 
Since gradients of masked tokens cannot be back-propagated, the position embeddings corresponding to padding tokens will not be updated. Thus, embedding vectors of front positions can be updated more times than the vectors of rear positions. We wonder whether the insufficient fine-tuning can make PLM have an inferior ability to represent rear tokens. 

As a pilot experiment, we train a BERT-base model on the \textit{Wikipedia} domain of TriviaQA~\citep{joshi-etal-2017-triviaqa}. 
We record how many times a position embedding has been updated during the fine-tuning process. 
We find that the first position embedding (i.e., for position No. 0) is updated in every step, however, the last one (i.e., for position No. 511) are updated in only 62.58\% of all steps. We further examine the differences in model performance when predicting answers at different positions. To be concise, we mainly focus on the predicted start boundary $P_s$ and regard it as the position of whole answer. 

As shown in Table~\ref{tab1_ans_pos}, the model can achieve a 65.36\% F1 to find answers appearing in the first quarter of positions, while it gets only a 57.03\% F1 in the last quarter. It shows the model tends to make more mistakes when predicting answer at the rear positions of the token sequence. From the first quarter to the last, we find the average F1 score over five runs keeps decreasing. Although this is not a strictly fair comparison, it still shows that QA models tend to provide more incorrect predictions on the tokens whose position embeddings have been fine-tuned fewer times. 

\begin{table}[t]
\centering
\begin{small}
\begin{tabular}{cc}
\hline
Answer Position & Average F1\\
\hline
$0\leq P_s <128$ & $65.36\pm 0.49$ \\
$128\leq P_s <256$ & $58.45\pm 0.74$ \\
$256\leq P_s <384$ & $57.42\pm 0.63$ \\
$384\leq P_s <512$ & $57.03\pm 0.51$ \\
\hline
\end{tabular}
\end{small}
\caption{\label{tab1_ans_pos}
Average F1 scores of the sub-sets divided by answer position.
}
\vskip -0.15in
\end{table}

\section{Our Method: Random Padding}
\label{sec:sec3_method}
As shown in our pilot study, 
when we train a model on instances of short contexts 
, embeddings at the front positions can be updated much more times than those at rear positions. Therefore, it is intuitive to balance the updating times over the whole range of positions, i.e., to reduce updating times of front position embeddings and to reallocate more updating to rear ones.

\begin{figure}[t]
\begin{center}
\centerline{\includegraphics[width=\columnwidth]{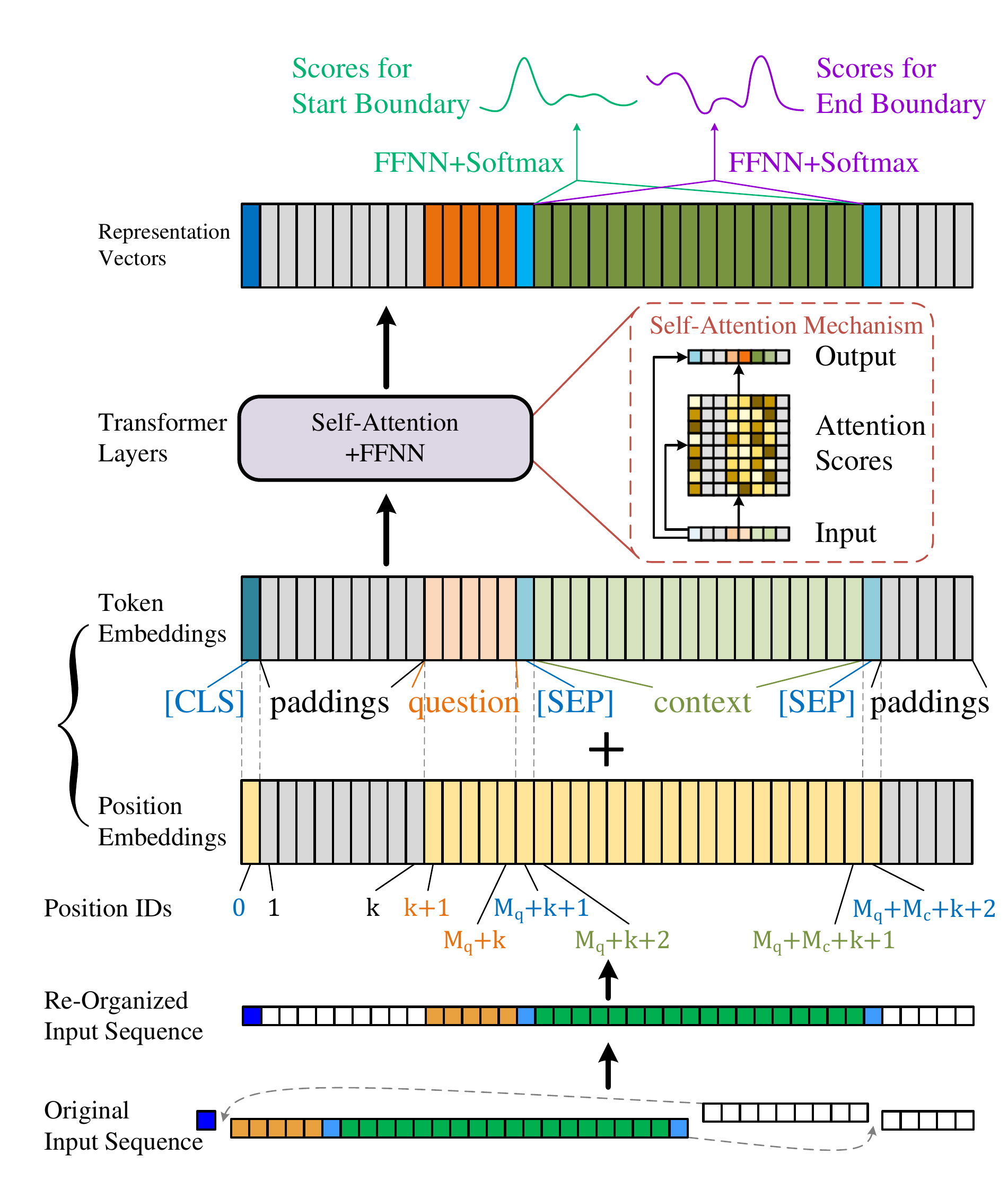}}
\caption{The process of QA model enhanced by Random Padding. }
\label{fig:QA_model_with_RP} 
\end{center}
\vskip -0.3in
\end{figure}

Recall that when fine-tuning a PLM for extractive QA, we only update the position embeddings of non-padding tokens. Since padding tokens are always at the rear positions of whole input sequence, these rear position embeddings are often ignored in the scheme of absolute position embedding. If padding tokens can be randomly placed in the whole sequence during fine-tuning, we can expect that every position embedding has almost equal chance to be updated or ignored. However, if we insert padding tokens into question or context, it will change the positional relationships of non-padding tokens, which might hurt model performance. Therefore, we should preserve the question tokens and the context tokens as contiguous sequences. Specially, during fine-tuning, we propose to move a random number of padding tokens to the front of the input sequence, as shown in Figure \ref{fig:QA_model_with_RP}. Then non-padding tokens will be pushed towards the end of input sequence, so that the rear position embeddings can be updated.


\paragraph{At the Fine-tuning Stage} Similar to Section \ref{sec:inv_model}, we  assume the upper limit of input length is $n$ tokens. For an input sequence with $m$ non-padding tokens (including \texttt{[CLS]}, \texttt{[SEP]}, \textit{question} and \textit{context}), there should exist $(n-m)$ padding tokens at the end. Then, we randomly select an integer $k\in\left[0,\ m-n\right]$, and move $k$ padding tokens from the rear to the front of the input sequence. The resulting input sequence will be:
\begin{equation}
    \begin{aligned}
        & \texttt{[CLS]}~\{\boldsymbol{k}\texttt{[PAD]}\}~\textit{question}~\texttt{[SEP]}~... \\ 
        & ...~\textit{context}~\texttt{[SEP]}~\{\boldsymbol{(m-n-k)}\texttt{[PAD]}\} \nonumber   
    \end{aligned}
\end{equation}

Note that we keep \texttt{[CLS]} at the $0$-th position, as the first token of input sequence. 
The reason is, under the widely adopted setting of SQuAD 2.0~\citep{rajpurkar-etal-2018-know}, QA models use the scores of \texttt{[CLS]} to decide whether it can find answers for given questions. 
Therefore, we believe 
\texttt{[CLS]} should be arranged at a fixed position during fine-tuning, to reduce interference and to accelerate convergence. 



\paragraph{At the Inference Stage} 
We do not move padding tokens during inference, and just feed the  original input sequence to the trained model:  
\begin{equation}
    \begin{aligned}
        & \texttt{[CLS]}~\textit{question}~\texttt{[SEP]}~... \\ 
        & ...~\textit{context}~\texttt{[SEP]}~\{\boldsymbol{(m-n)}\texttt{[PAD]}\} \nonumber 
    \end{aligned}
\end{equation}
Note that the input sequence during inference will be the same as baseline method.

To summarize, we simply adjust the token order of input sequence when fine-tuning, by moving a random number of padding tokens to the position between \texttt{[CLS]} and \textit{question} tokens. Our method does NOT make any modifications to the architectures of existing PLMs, thus is convenient to be employed by PLMs with absolute position embeddings.

\section{Experiments}

We experiment on three different PLMs to answer the following questions: (1)~When models are trained on short contexts but evaluated on long contexts, can Random Padding improve the performance of QA models? (2)~If models are trained and evaluated on datasets with similar distributions of context lengths, can Random Padding still work? (3)~Does Random Padding actually improve model performance more when answers are located at the end of contexts? (4)~How does Random Padding play a role to improve QA models?

\subsection{Dataset Preparation}
\label{sec:datasets_define}

Similar to Section \ref{sec:pilot_exp}, we experiment on the \textit{Wikipedia} domain of TriviaQA~\citep{joshi-etal-2017-triviaqa}. Since we cannot access the official test set, we randomly split the official development set into two equal sub-sets for validation and test. 

To better understand the behaviours of our method, we investigating the models with or without Random Padding on various training/testing conditions, e.g., training set and test set may have different distributions of context lengths. Specifically, we prepare two different kinds of datasets.

The first one is to truncate all long context to a fixed length, e.g., 100 or 800 words. 
We denote the datasets truncated to a fixed length $L$ as $\mathcal{D}_L$, and its corresponding training/validation/test sets as $\mathcal{D}_L^{train},\ \mathcal{D}_L^{val},\ \mathcal{D}_L^{test}$, repsectively.

The second type is to truncate long contexts to a range of lengths. We pre-define the minimum and maximum length limitations as $L_1$ and $L_2$. For each instance in original TriviaQA$_{\textit{Wiki}}$ dataset, we randomly select an integer $k\in\left[L_1,\ L_2\right]$ and truncate its context to $k$ words. Similarly, its training/validation/test sets are $\mathcal{D}_{L_1\sim L_2}^{train},\ \mathcal{D}_{L_1\sim L_2}^{val},\ \mathcal{D}_{L_1\sim L_2}^{test}$, respectively.

These two types of datasets correspond to two practical scenarios. For the automatically annotated datasets, like TriviaQA~\citep{joshi-etal-2017-triviaqa} and SearchQA~\citep{dunn-etal-2017-searchQA}, their contexts are gathered from multiple documents relevant to the questions, 
and truncated to a fixed length when creating the datasets. For the manually annotated datasets, like SQuAD v1 \& v2~\citep{rajpurkar-etal-2016-squad, rajpurkar-etal-2018-know}, human annotators write questions and answers for every single passage, where the contexts are not truncated and have various lengths.

\subsection{Implementation Details}
%
We investigate three different extractive QA models, with the \texttt{base} version of BERT~\citep{devlin-etal-2019-bert}, RoBERTa~\citep{liu-etal-2019-roberta}, and ELECTRA~\cite{Clark2020ELECTRA}, respectively. We fine-tune all models under baseline settings, which takes the original pre-processed sequences as input~(with the same form in Section~\ref{sec:inv_model}). For each model, we employ Adam~\citep{adam} as optimizer and run experiments five times with different seeds. For every run to train baseline models, we also record the shuffled orders of training instances in every epoch.


When fine-tuning models enhanced by Random Padding, we utilize the same hyper-parameters as baseline models.
In every epoch over five runs, we also keep the orders of training instances identical to baseline settings, to eliminate the stochastic influence of training orders on model performance.
More implement details and dataset statistics are shown in Appendix \ref{app:exp_detail}.

\section{Main Results} 
\label{sec:results_4_3}

\subsection{Train Short, Test Long}
\label{sec:train_short_test_long}

We first look at whether Random Padding can help models to obtain better embeddings of rear positions, when there do not exist long contexts in the training set. Specifically, all models are trained on instances with short context only~(e.g., $\mathcal{D}_{100}^{train}$ defined in Section \ref{sec:datasets_define}), but have to be evaluated on instances with longer contexts~(e.g., $\mathcal{D}_{100\sim 800}^{val}$ and $\mathcal{D}_{100\sim 800}^{test}$). It is a practical and hard scenario, where training data and test data are sampled from different distributions.


\begin{table}[t]
\centering
\tabcolsep=0.27em
\begin{small}
\begin{tabular}{clcccc}
\toprule
\multirow{2}*{\makecell{Training\\Set}} & \multirow{2}*{Model} & \multicolumn{2}{c}{Validation} & \multicolumn{2}{c}{Test} \\
\cmidrule(lr){3-4} \cmidrule(lr){5-6}
 & & F1 & EM & F1 & EM \\
\midrule
& BERT & 58.78$_{\text{0.25}}$ & 52.17$_{\text{0.51}}$ & 58.75$_{\text{0.45}}$ & 51.87$_{\text{0.47}}$ \\
& +RP & \textbf{59.86}$_{\text{0.47}}$ & \textbf{54.12}$_{\text{0.40}}$ & \textbf{59.82}$_{\text{0.40}}$ & \textbf{53.60}$_{\text{0.48}}$ \\
\cmidrule{2-6}
\multirow{2}*{$\mathcal{D}_{100}^{train}$} &ELEC & 61.67$_{\text{0.21}}$ & 55.54$_{\text{0.08}}$ & 61.31$_{\text{0.15}}$ & 54.80$_{\text{0.16}}$ \\
&+RP & \textbf{63.32}$_{\text{0.35}}$ & \textbf{57.16}$_{\text{0.62}}$ & \textbf{62.76}$_{\text{0.44}}$ & \textbf{56.45}$_{\text{0.63}}$ \\
\cmidrule{2-6}
&RoBE & 61.74$_{\text{0.17}}$ & 55.39$_{\text{0.35}}$ & 61.07$_{\text{0.43}}$ & 54.44$_{\text{0.62}}$ \\
&+RP & \textbf{62.98}$_{\text{0.16}}$ & \textbf{56.73}$_{\text{0.32}}$ & \textbf{62.44}$_{\text{0.35}}$ & \textbf{55.94}$_{\text{0.58}}$ \\
\midrule
& BERT & 62.61$_{\text{0.13}}$ & 57.02$_{\text{0.32}}$ & 62.91$_{\text{0.15}}$ & 56.94$_{\text{0.27}}$ \\
&+RP & \textbf{63.46}$_{\text{0.32}}$ & \textbf{57.84}$_{\text{0.51}}$ & \textbf{63.97}$_{\text{0.40}}$ & \textbf{58.01}$_{\text{0.49}}$ \\
\cmidrule{2-6}
\multirow{2}*{$\mathcal{D}_{250}^{train}$} & ELEC & 66.16$_{\text{0.28}}$ & 60.66$_{\text{0.57}}$ & 66.14$_{\text{0.28}}$ & 60.30$_{\text{0.55}}$ \\
&+RP & \textbf{66.88}$_{\text{0.14}}$ & \textbf{61.46}$_{\text{0.25}}$ & \textbf{66.94}$_{\text{0.27}}$ & \textbf{61.41}$_{\text{0.34}}$ \\
\cmidrule{2-6}
&RoBE & 65.11$_{\text{0.20}}$ & 59.44$_{\text{0.39}}$ & 65.01$_{\text{0.40}}$ & 59.00$_{\text{0.37}}$ \\
&+RP & \textbf{65.60}$_{\text{0.20}}$ & \textbf{60.01}$_{\text{0.12}}$ & \textbf{65.47}$_{\text{0.33}}$ & \textbf{59.63}$_{\text{0.40}}$ \\
\midrule
& BERT & 64.03$_{\text{0.14}}$ & 58.48$_{\text{0.46}}$ & 64.79$_{\text{0.29}}$ & 58.96$_{\text{0.37}}$ \\
&+RP & \textbf{64.90}$_{\text{0.27}}$ & \textbf{59.55}$_{\text{0.35}}$ & \textbf{65.21}$_{\text{0.33}}$ & \textbf{59.55}$_{\text{0.34}}$ \\
\cmidrule{2-6}
\multirow{2}*{$\mathcal{D}_{400}^{train}$} & ELEC & 67.76$_{\text{0.35}}$ & 62.61$_{\text{0.40}}$ & \textbf{68.07}$_{\text{0.34}}$ & 62.28$_{\text{0.44}}$ \\
&+RP & \textbf{68.09}$_{\text{0.14}}$ & \textbf{62.77}$_{\text{0.25}}$ & 67.94$_{\text{0.21}}$ & \textbf{62.37}$_{\text{0.23}}$ \\
\cmidrule{2-6}
&RoBE & 66.67$_{\text{0.16}}$ & 61.25$_{\text{0.19}}$ & 66.34$_{\text{0.26}}$ & 60.75$_{\text{0.29}}$ \\
&+RP & \textbf{66.82}$_{\text{0.09}}$ & \textbf{61.38}$_{\text{0.16}}$ & \textbf{66.78}$_{\text{0.30}}$ & \textbf{61.26}$_{\text{0.27}}$ \\
\bottomrule
\end{tabular}
\caption{\label{tab3_short_train}
Performance of models which are trained on short contexts but evaluated on long contexts~($\mathcal{D}^{val}_{100\sim 800}$ and $\mathcal{D}^{test}_{100\sim 800}$). ELECTRA and RoBERTa are abbreviated as \textit{ELEC} and \textit{RoBE}. 
}
\end{small}
\vskip -0.15in
\end{table}


As shown in Table \ref{tab3_short_train}, when models are trained on the contexts with around 100 words~($\mathcal{D}_{100}^{train}$), Random Padding can bring an improvement more than +1\% F1 on the test set, specifically +1.07\% for BERT, +1.45\% for ELECTRA, and +1.37\% for RoBERTa. 
When we employ $\mathcal{D}_{250}^{train}$ as the training set, Random Padding can only bring an improvement of +1.06\% / +0.80\% / +0.46\% F1 for BERT / ELECTRA / RoBERTa, respectively. It is not surprising that models trained on shorter contexts~($\mathcal{D}_{100}^{train}$ v.s. $\mathcal{D}_{250}^{train}$ or $\mathcal{D}_{400}^{train}$) can gain more improvement from Random Padding, since more position embeddings at rear positions cannot be updated when models are traiend on shorter contexts.

\subsection{Train/Test with Similar Context Lengths}


We also wonder whether Random Padding can help all three PLMs to gain improvement when training set and test set have similar distributions of context lengths. Here, we will use the second type of datasets created in Section~\ref{sec:datasets_define}. For example, 
all models will be trained on $\mathcal{D}_{800}^{train}$, validated and tested on $\mathcal{D}_{800}^{val}$ and $\mathcal{D}_{800}^{test}$, respectively.
 
%
Table \ref{tab2_homogeneous} shows the performance of three PLMs with and without our Random Padding strategy. We can see that Random Padding brings +0.73\% F1 improvement for BERT on $\mathcal{D}^{test}_{100\sim 800}$, +0.29\% for ELECTRA and +0.32\% for RoBERTa. When trained and tested on $\mathcal{D}_{800}$, models can also gain improvement from Random Padding, specifically +0.38\% F1 for BERT, +0.22\% F1 for ELECTRA, and +0.26\% for RoBERTa. 

We find all three PLMs gain more improvement from Random Padding on $\mathcal{D}_{100\sim 800}^{test}$ than $\mathcal{D}_{800}^{test}$. It is similar to the results in \ref{sec:train_short_test_long}, showing that if the training set consists of more instances with short contexts, Random Padding will be more effective to improve model performance. Comparing the results of three PLMs on the same dataset, we also find Random Padding boost the performance of BERT more than RoBERTa and ELECTRA. It might because BERT are pre-trained with sequence length of 128 for 90\% of the steps, while ELECTRA and RoBERTa are pre-trained on sequences with 512 tokens for all steps. Thus, ELECTRA and RoBERTa obtain better position embeddings from pre-training than BERT. Moreover, ELECRTA and RoBERTa are pre-trained with larger batch size and on larger corpora, which enhances their ability to generate contextual representations, which, we guess, leaves less space for Random Padding to improve the performance  of ELECTRA and RoBERTa.


\begin{table}[t]
\centering
\tabcolsep=0.42em
\begin{small}
\subfigure[Model performance on $\mathcal{D}_{800}$]{
\begin{tabular}{lcccc}
\toprule
\multirow{2}*{Model} & \multicolumn{2}{c}{Validation} & \multicolumn{2}{c}{Test} \\
\cmidrule(lr){2-3} \cmidrule(lr){4-5}
& F1 & EM & F1 & EM \\
\midrule
BERT & 60.74$_{\text{0.28}}$ & 54.48$_{\text{0.33}}$ & 61.45$_{\text{0.23}}$ & 55.54$_{\text{0.26}}$\\
+RP & \textbf{61.18}$_{\text{0.09}}$ & \textbf{54.92}$_{\text{0.30}}$ & \textbf{61.83}$_{\text{0.33}}$ & \textbf{55.98}$_{\text{0.35}}$\\
\cmidrule{1-5}
ELECTRA & 64.56$_{\text{0.29}}$ & 58.92$_{\text{0.28}}$ & 65.48$_{\text{0.23}}$ & 59.59$_{\text{0.14}}$\\
+RP & \textbf{64.77}$_{\text{0.17}}$ & \textbf{59.07}$_{\text{0.25}}$ & \textbf{65.70}$_{\text{0.26}}$ & \textbf{59.81}$_{\text{0.38}}$\\
\cmidrule{1-5}
RoBERTa & 63.11$_{\text{0.26}}$ & 56.87$_{\text{0.36}}$ & 63.24$_{\text{0.14}}$ & 57.23$_{\text{0.22}}$ \\
+RP & \textbf{63.24}$_{\text{0.17}}$ & \textbf{57.08}$_{\text{0.14}}$ & \textbf{63.50}$_{\text{0.30}}$ & \textbf{57.44}$_{\text{0.27}}$ \\
\bottomrule
\end{tabular}
}

\subfigure[Model performance on $\mathcal{D}_{100\sim 800}$]{
\begin{tabular}{lcccc}
\toprule
\multirow{2}*{Model} & \multicolumn{2}{c}{Validation} & \multicolumn{2}{c}{Test} \\
\cmidrule(lr){2-3} \cmidrule(lr){4-5}
& F1 & EM & F1 & EM \\
\midrule
BERT & 63.65$_{\text{0.16}}$ & 58.11$_{\text{0.34}}$ & 64.33$_{\text{0.41}}$ & 58.36$_{\text{0.39}}$\\
+RP & \textbf{64.40}$_{\text{0.08}}$ & \textbf{58.97}$_{\text{0.16}}$ & \textbf{65.06}$_{\text{0.29}}$ & \textbf{59.28}$_{\text{0.24}}$\\
\cmidrule{1-5}
ELECTRA & 67.14$_{\text{0.18}}$ & 61.75$_{\text{0.23}}$ & 67.21$_{\text{0.20}}$ & 61.66$_{\text{0.24}}$\\
+RP & \textbf{67.35}$_{\text{0.13}}$ & \textbf{61.81}$_{\text{0.25}}$ & \textbf{67.50}$_{\text{0.14}}$ & \textbf{61.84}$_{\text{\text{0.26}}}$\\
\cmidrule{1-5}
RoBERTa & 65.77$_{\text{0.21}}$ & 60.22$_{\text{0.38}}$ & 66.05$_{\text{0.15}}$ & 60.21$_{\text{0.16}}$ \\
+RP & \textbf{66.11}$_{\text{0.28}}$ & \textbf{60.51}$_{\text{0.40}}$ & \textbf{66.37}$_{\text{0.19}}$ & \textbf{60.50}$_{\text{0.35}}$ \\
\bottomrule
\end{tabular}
}
\end{small}
\caption{\label{tab2_homogeneous}
Model performance on the datasets whose training set and test set have similar distributions of context lengths. 
}
\vskip -0.15in
\end{table}

\section{Analysis and Discussions}
\label{sec:sec6_discussion}

Experimental results reveal that Random Padding can effectively improve extractive QA models, especially when the models are trained on short contexts but evaluated on long contexts~(Table \ref{tab3_short_train}). Considering Random Padding can make rear position embeddings updated more times, we wonder whether it brings more improvement when models predict answers at the rear part of context. We also wonder how Random Padding improves QA models.

\begin{table}[t]
\centering
\begin{small}

\subfigure[Results on $\mathcal{V}$~(the validation set of $\mathcal{D}_{100\sim 800}$)]{
\begin{tabular}{ccccc}
\toprule
Training & \multirow{2}*{Seg.} & \multicolumn{2}{c}{F1 Scores} & \multirow{2}*{$\Delta$F1}\\
\cmidrule(lr){3-4}
Set & & BERT & BERT+RP & \\
\midrule
\multirow{2}*{$\mathcal{D}_{100}^{train}$} & $\mathcal{V}_{100}^{(1)}$ & 72.03±0.55 & 71.88±0.47 & -0.15 \\
& $\mathcal{V}_{100}^{(2)}$ & 38.97±1.00 & 41.88±0.72 & \textbf{+2.91} \\ 
\midrule
\multirow{2}*{$\mathcal{D}_{250}^{train}$} & $\mathcal{V}_{250}^{(1)}$ & 68.76±0.24 & 69.32±0.47 & +0.56 \\
& $\mathcal{V}_{250}^{(2)}$ & 36.65±0.65 & 38.63±1.06 & \textbf{+1.98} \\ 
\midrule
\multirow{2}*{$\mathcal{D}_{400}^{train}$} & $\mathcal{V}_{400}^{(1)}$ & 67.03±0.19 & 67.86±0.35 & +0.83 \\
& $\mathcal{V}_{400}^{(2)}$ & 37.19±0.99 & 38.39±0.91 & \textbf{+1.21} \\ 
\bottomrule
\end{tabular}
}

\subfigure[Results on $\mathcal{T}$~(the test set of $\mathcal{D}_{100\sim 800}$)]{
\begin{tabular}{ccccc}
\toprule
Training & \multirow{2}*{Seg.} & \multicolumn{2}{c}{F1 Scores} & \multirow{2}*{$\Delta$F1}\\
\cmidrule(lr){3-4}
Set & & BERT & BERT+RP & \\
\midrule
\multirow{2}*{$\mathcal{D}_{100}^{train}$} & $\mathcal{T}_{100}^{(1)}$ & 71.92±0.85 & 71.14±0.54 & -0.78 \\
& $\mathcal{T}_{100}^{(2)}$ & 39.67±0.57 & 43.43±0.71 & \textbf{+3.76} \\ 
\midrule
\multirow{2}*{$\mathcal{D}_{250}^{train}$} & $\mathcal{T}_{250}^{(1)}$ & 68.53±0.15 & 69.48±0.35 & +0.95 \\
& $\mathcal{T}_{250}^{(2)}$ & 39.60±1.00 & 41.24±1.35 & \textbf{+1.64} \\ 
\midrule
\multirow{2}*{$\mathcal{D}_{400}^{train}$} & $\mathcal{T}_{400}^{(1)}$ & 67.28±0.50 & 67.71±0.32 & \textbf{+0.43} \\
& $\mathcal{T}_{400}^{(2)}$ & 41.92±1.32 & 42.20±2.02 & +0.28 \\
\bottomrule
\end{tabular}
}
\caption{\label{tab4_ablation_position_valset}
Improvement~($\Delta$F1) from Random Padding on every Segment of $\mathcal{D}_{100\sim 800}^{val}$ and $\mathcal{D}_{100\sim 800}^{test}$. 
}
\end{small}
\vskip -0.15in
\end{table}

\subsection{Analysis on Answer Positions}

To study the relationship between improvement from Random Padding and answer positions, we divide test set $\mathcal{S}$ into two segments, $\mathcal{S}^{(1)}_{X}$  and $\mathcal{S}^{(2)}_{X}$. The first segment, $\mathcal{S}^{(1)}_{X}$ , contains the instances, at least one of whose answers is located in the front $X$ tokens of the input sequence, while all other instances are gathered into the second segment, $\mathcal{S}^{(2)}_{X}$. We provide detailed explanations about our segment division in Appendix \ref{app:seg_division}.

We train a QA model with BERT on $\mathcal{D}^{train}_{X}$, whose contexts are truncated to around $X$ words~(100, 250, or 400). We then split the validation set and test set into two segments according to $X$, to examine whether models with or without Random Padding perform differently regarding answer positions. Here, we employ $\mathcal{D}_{100\sim 800}^{val}$ and $\mathcal{D}_{100\sim 800}^{test}$ as our validation and test sets. To be simplified, we denote them as $\mathcal{V}$ and $\mathcal{T}$ in Table \ref{tab4_ablation_position_valset}, which shows the improvement brought by Random Padding on each segment.


We find Random Padding brings significant improvement on the \textit{Second Segment}, for example, +3.76\% F1 improvement on $\mathcal{T}_{100}^{(2)}$ when trained on $\mathcal{D}_{100}^{train}$. However, we surprisingly find that models can also obtain no more than +1\% F1 improvement for the \textit{First Segment}. When trained on $\mathcal{D}_{100}^{train}$, models even perform a little worse on the \textit{First Segment} after enhanced by Random Padding. It reveals the main contribution of Random Padding is to improve model performance on the instances whose answers are all located at rear positions. 

We also notice when $X$ is 400, Random Padding brings less improvement on the \textit{Second Segment} of $\mathcal{D}^{test}_{100\sim 800}$ than the \textit{First Segment}. In $\mathcal{T}^{(1)}_{400}$ and $\mathcal{T}^{(2)}_{400}$, there exists 4,247 and 462 instances respectively, while the numbers for $\mathcal{T}^{(1)}_{100}$ and $\mathcal{T}^{(2)}_{100}$ are 2,786 and 1,923. 
It is not surprising that the volatility of scores might rise when evaluated on such small dataset as $\mathcal{T}_{400}^{(2)}$, which causes the results of $X=400$ to become an outlier.


\begin{table}[t]
\centering
\begin{small}
\begin{tabular}{cclcc}
\toprule
\makecell{Seg-\\ment} & \makecell{Training\\Set} & Model & \makecell{F1 Score}& $\Delta$F1\\
\midrule
\multirow{6}*{$\mathcal{V}_{100}^{(2)}$} & \multirow{2}*{$\mathcal{D}_{100}^{train}$} & BERT & 38.97±1.00 & \multirow{2}*{\textbf{2.91}}\\
 & & +RP & 41.88±0.72 & \\
\cmidrule(l){2-5}
& \multirow{2}*{$\mathcal{D}_{250}^{train}$} & BERT & 45.78±0.46 & \multirow{2}*{1.75}\\
 & & +RP & 47.53±0.77 & \\
\cmidrule(l){2-5}
& \multirow{2}*{$\mathcal{D}_{400}^{train}$} & BERT & 49.52±0.46 & \multirow{2}*{1.68}\\
 & & +RP & 51.20±0.40 & \\
\midrule\multirow{6}*{$\mathcal{T}_{100}^{(2)}$} & \multirow{2}*{$\mathcal{D}_{100}^{train}$} & BERT & 39.67±0.57 & \multirow{2}*{\textbf{3.76}}\\
 & & +RP & 43.43±0.71 & \\
\cmidrule(l){2-5}
& \multirow{2}*{$\mathcal{D}_{250}^{train}$} & BERT & 47.38±0.86 & \multirow{2}*{1.72}\\
 & & +RP & 49.10±0.70 & \\
\cmidrule(l){2-5}
& \multirow{2}*{$\mathcal{D}_{400}^{train}$} & BERT & 51.51±1.13 & \multirow{2}*{0.93}\\
 & & +RP & 52.44±0.66 & \\
\bottomrule
\end{tabular}
\caption{\label{tab6_fixed_segments}
Performance on fixed division of segments.
}
\end{small}
\vskip -0.15in
\end{table}

\paragraph{Random Padding v.s. Longer Context} The results in Table \ref{tab3_short_train} and \ref{tab4_ablation_position_valset} indicate that models trained on longer contexts will gain less improvement from Random Padding. Considering the main contribution of Random Padding mentioned above, we are interested to see whether the improvement on the \textit{Second Segment} will decrease if models are trained on longer contexts. Here, we train BERT models on instances with different context lengths ($\mathcal{D}_{100}^{train}$, $\mathcal{D}_{250}^{train}$, or $\mathcal{D}_{400}^{train}$), and then evaluate the models on a fixed division of segments ($\mathcal{V}^{(2)}_{100}$ and $\mathcal{T}^{(2)}_{100}$).


The results are shown in Table \ref{tab6_fixed_segments}. For the model trained on $\mathcal{D}_{100}^{train}$, Random Padding can bring +2.91\% F1 and +3.76\% F1 improvement on $\mathcal{V}^{(2)}_{100}$ and $\mathcal{T}^{(2)}_{100}$. When we use training set with longer contexts, the improvement decreases to +0.93\%$\sim$+1.75\% F1. It indicates the models trained on longer contexts will gain less improvement when predicting answers at rear positions. One reason may be that longer contexts can also make rear position embeddings to be updated more times, which plays similar role as Random Padding.


\subsection{How Random Padding Improves QA Performance?} %


To study how Random Padding works, we consider the improvement on a specific instance in two categories: (1) Baseline model fails to find the right answer, while our improved model can make corrections to wrong predictions. (2) Baseline model has almost found the right answer, but it selects redundancy tokens around the ground-truth, which are then filtered out by our improved model. We provide example cases in Table \ref{tab:example_case}.

\begin{table}[h]
\centering
\begin{footnotesize}
\begin{tabular}{p{0.12\columnwidth}|p{0.75\columnwidth}}
\hline
\makecell[l]{Type} & \makecell{Example Case}\\
\Xhline{0.4pt}
\multirow{2}{*}[-4ex]{\makecell[l]{Make\\Correc-\\tions}} & \casefont{Question: In the TV series Thunderbirds, Parker was chauffeur to whom?}\\
& \casefont{Context: Parker ... appears in the film sequels \underline{Thunderbirds Are Go} (1966) ... Parker is employed at Creighton-Ward Mansion by \red{\sethlcolor{yellow}\hl{Lady Penelope}}, serving as her butler and chauffeur... }\\
\hline
\multirow{2}{*}[-2.5ex]{\makecell[l]{Find\\Precise\\Boun-\\daries}} & \casefont{Question: Which metal is produced by the Bessemer Process?}\\
& \casefont{Context: The Bessemer process was the first inexpensive industrial process for the mass-production of \underline{\red{\sethlcolor{yellow}\hl{steel}} from molten pig iron} prior to the open hearth furnace ...}\\
\hline
\end{tabular}
\caption{\label{tab:example_case}
Different types of examples corrected by our method. The \underline{underlined texts} are wrong predictions given by baseline models. \red{Red texts} are correct predictions given by the improved model. \sethlcolor{yellow}\hl{Texts with yellow background} are golden answer.
}
\end{footnotesize}
\end{table}

\begin{table*}[t]
\centering
\begin{small}
\begin{tabular}{lcccccc}
\toprule
\multirow{2}*{Model} & \multicolumn{2}{c}{Natural Questions} & \multicolumn{2}{c}{HotpotQA} & \multicolumn{2}{c}{SQuAD 2.0} \\
\cmidrule(lr){2-3} \cmidrule(lr){4-5} \cmidrule(lr){6-7}
 & Eval F1 & Test F1 & Eval F1 & Test F1 & Eval F1 & Test F1 \\
\midrule
BERT & 78.49±0.20 & 78.18±0.24 & 75.46±0.12 & 73.41±0.29 & 77.69±0.50 & 77.71±0.55\\
+RP & \textbf{78.86}±0.35 & \textbf{78.67}±0.33 & \textbf{76.22}±0.34 & \textbf{74.23}±0.15 & \textbf{78.04}±0.41 & \textbf{78.00}±0.47\\
\midrule
ELECTRA & 81.30±0.17 & 80.71±0.18 & 78.59±0.25 & 76.55±0.13 & \textbf{84.44}±0.38 & 83.85±0.34\\
+RP & \textbf{81.46}±0.13 & \textbf{80.88}±0.13 & \textbf{78.71}±0.29 & \textbf{76.67}±0.28 & 84.40±0.55 & \textbf{83.95}±0.34\\
\midrule
RoBERTa & 80.67±0.15 & 80.43±0.18 & 79.63±0.20 & 76.87±0.28 & 83.03±0.41 & 82.85±0.26\\
+RP & \textbf{80.81}±0.22 & \textbf{80.74}±0.27 & \textbf{79.73}±0.41 & \textbf{77.42}±0.26 & \textbf{83.37}±0.14 & \textbf{82.99}±0.18\\
\bottomrule
\end{tabular}
\end{small}
\caption{\label{tab8_public}
Performance on public extractive QA benchmark datasets.
}
\vskip -0.15in
\end{table*}

For every instance, we compare the predictions provided by baseline and improved models. If the prediction of baseline model obtains a lower F1 score than improved model, we will
compare the two predictions. If there exists no common sub-sequence between the two sequences, we will consider improved model selects the correct answer but baseline model not. Otherwise, the improved model will be regarded to find a more precise answer boundary.


We take the BERT model trained on $\mathcal{D}_{100}^{train}$ as baseline. We then enhance the model by Random Padding, or longer contexts ($\mathcal{D}_{400}^{train}$). All models are trained with five different random seeds and tested on $\mathcal{D}^{test}_{100\sim 800}$. Among the instances improved by Random Padding, 36.11\%±2.75\% of them are improved by finding more precise boundaries, while the rest are because of making corrections to wrong answers. However, for the instances improved by training with longer contexts, 31.73\%±2.59\% of them can be attributed to more precise boundaries. From the results, we think  that Random Padding can enhance the ability of QA models to better represent the positional information between neighbour tokens, which leads more improvement on answer boundaries.

\begin{table}[h]
\centering
\begin{small}
\begin{tabular}{lcc}
\toprule
Model & Test F1 & Test Ign F1 \\
\midrule
BERT-base & 73.57±0.20 & 72.74±0.19\\
BERT-base+RP & \textbf{73.92}±0.09 & \textbf{73.10}±0.14\\
\midrule
BERT-large & 76.06±0.11 & 75.40±0.09\\
BERT-large+RP & \textbf{76.18}±0.10 & \textbf{75.53}±0.11\\
\midrule
RoBERTa-base & 75.29±0.19 & 74.57±0.20\\
RoBERTa-base+RP & \textbf{75.42}±0.23 & \textbf{74.71}±0.31\\
\midrule
RoBERTa-large & 77.78±0.24 & 77.10±0.25\\
RoBERTa-large+RP & \textbf{78.00}±0.25 & \textbf{77.34}±0.27\\
\bottomrule
\end{tabular}
\end{small}
\caption{\label{tab9_docred}
Model performance on Re-DocRED.
}
\vskip -0.15in
\end{table}

\section{Results on More Benchmark Datasets}
\label{sec:sec7_openbenchmark}


Now we proceed to examine the effect of Random Padding on more benchmark datasets. We train and evaluate QA models with various PLMs on NaturalQuestions~\citep{Kwiatkowski-et-al-19-NaturalQ}, HotpotQA~\citep{yang-etal-2018-hotpotqa}, and SQuAD 2.0~\cite{rajpurkar-etal-2018-know}. Since these benchmarks provide validation set merely, we randomly split the original validation sets to two parts as our validation and test sets. 


As shown in Table~\ref{tab8_public}, 
we can observe our simple Random Padding helps BERT obtain +0.82\% F1 improvement on HotpotQA, +0.49\% F1 on NaturalQuestions, and +0.29\% F1 on SQuAD v2, while ELECRTA and RoBERTa can only gain improvement of around +0.3\% F1 or even no significant improvement. It is similar to the results in Section \ref{sec:results_4_3}. Since ELECTRA and RoBERTa have been trained with larger batch size and larger corpora, there is less space left to Random Padding to improve.

Besides extractive question answering, we also study the document-level relation extraction~(RE) task, where models should provide high-quality representations to encode local context~\citep{zhang-et-al-2021-document,wang-etal-2020-global}. As discussed in Section~\ref{sec:sec6_discussion}, Random Padding can enhance model's ability to encode relative positions of neighbour tokens, which provides important information in representing local context. Here, we employ the state-of-the-art RE framework, ATLOP~\citep{zhou2021atlop}, and examine both BERT and RoBERTa as the PLM component in ATLOP. Table \ref{tab9_docred} shows the  model performance on the Re-DocRED~\citep{tan2022revisiting} benchmark. We find BERT can gain improvement of +0.35\% F1 and +0.36\% Ign F1 from Random Padding. The results also show Random Padding improves BERT more than RoBERTa, and improves \texttt{base} models more than \texttt{large} models. 






\section{Conclusion}

In this work, we propose a simple strategy, Random Padding, to improve the performance of extractive QA models, especially when they are trained on short contexts but evaluated on longer contexts. Our method only re-organizes the input token sequences when fine-tuning, without any modifications to the architectures of PLMs. Experiments reveal that our simple method can effectively enhance QA models when predicting answers at the rear positions, where the position embeddings may not be sufficiently updated without Random Padding. We also show that our simple strategy can improve the performance of PLM components in more benchmarks and tasks where accurate local context representations over longer context  are necessary. 

\section*{Limitations}

In this paper, we mainly focus on whether and how Random Padding improve the models for extractive question answering. We have to note that Random Padding cannot improve Transformer-based models with absolute position embeddings on all tasks. For example, we experiment on the tasks with a form of sentence classification, like Natural Language Inference. We train a text-classification model with BERT on MNLI~\citep{williams-etal-2018-broad}. Since the lengths of most texts in MNLI are not longer than 100 words, there is no need to update the position embeddings behind the 100-th position. Thus, we limit the maximum number of moved padding tokens to 32 during fine-tuning. Without Random Padding, the vanilla BERT achieves 92.69\%±0.09\% of accuracy, while the model with Random Padding achieves 92.61\%±0.23\%. We find that Random Padding can bring almost no influence to BERT on MNLI. Since MNLI is a text classification task, we should input the representation of \texttt{[CLS]} token to the class decoder, without any other representations. The representation of \texttt{[CLS]} should capture global contextual information of the whole text, while the main contribution of Random Padding is to enhance the representing ability of models for local context. Therefore, Random Padding cannot be effective for text classification. 

We also investigate whether Random Padding can be effective when models are train on instances with very long contexts. Here, we take NewsQA~\citep{trischler-etal-2017-newsqa} as an example. More 30\% of the instances in NewsQA have a context with no less than 1,000 words. Since BERT can only receive an input sequence of 512 tokens, these extremely long context will be splitted into multiple sequences. Therefore, most of the training sequences will contain 512 non-padding tokens, and every position embedding can be properly fine-tuned. We find the vanilla BERT model achieve 70.96\%±0.22\% F1 on NewsQA, while BERT enhanced by Random Padding can achieve 70.93\%±0.37\% F1. It show Random Padding has no contributions to the QA model trained on extremely long contexts. since there exist almost no padding tokens to be moved when fine-tuning.

As a summary, we note that Random Padding should be utilized in the tasks which need models to capture local context features, but not global features. And Random Padding cannot play a role to improve the models trained on extreme long contexts, whose position embeddings have all been properly fine-tuned. 




\bibliographystyle{acl_natbib}
\bibliography{anthology.bib,custom.bib}

\appendix

\section{Settings of Pilot Experiments}
\label{app:pilot_settings}

In the pilot experiments, we investigate the vanilla BERT-base model trained on the \textit{Wikipedia} domain of TriviaQA~\citep{joshi-etal-2017-triviaqa}. Similar to MRQA~\citep{fisch2019mrqa}, an open question answering benchmark, we convert the original dataset to SQuAD format~\citep{rajpurkar-etal-2016-squad} and truncate long documents to 800 words, via the scripts\footnote{\url{https://github.com/mandarjoshi90/triviaqa/blob/master/utils/convert_to_squad_format.py}} provided by \citet{joshi-etal-2017-triviaqa}. Since BERT can receive input sequences no longer than 512, a long context will be split into several sequences. Therefore, every instance in the training set will be divided into multiple tokenized sequences. And the last sequence will be much shorter than 512 tokens. Thus, when fine-tuning, the first position embedding is updated in every step, but the last one can be updated in only $62.58\%$ of all steps. We fine-tune the QA model based on BERT-base for 4 epochs, with batch size of 16 and learning rate of $3\times10^{-5}$.

We evaluate the model on validation set and collect the predicted answer position of every instance. Then, these instances are divided into four sub-sets via answer positions, and we calculate the average F1 scores of every sub-set respectively.

Moreover, it is worth noting that biases in answer position can also affect the performances of models~\citep{ko-etal-2020-look,shinoda-etal-2022-look}. For example, if all answers in training set are annotated in the first sentence of every context, QA models will also learn to give predictions in the first sentences. 

To eliminate position bias of answers, we re-annotate answer spans of the training set: For a given training instance, if the answer is mentioned in context more than one time, we will randomly select one from all mentions as the final annotation. Then, in training set, answers tend to be uniformly distributed over the whole context. 

\section{More Implement Details and Dataset Statistics}
\label{app:exp_detail}
\subsection{Implement Details}

In Section \ref{sec:results_4_3}, we train models on $\mathcal{D}_{800}^{train}$, $\mathcal{D}_{100\sim 800}^{train}$, $\mathcal{D}_{100}^{train}$, $\mathcal{D}_{250}^{train}$, or $\mathcal{D}_{400}^{train}$ with batch size of 16. Adam~\citep{adam} is employed as the optimizer. For each experiment, we evaluate model performance every 1000 steps on validation sets and save the best checkpoints for test. Each experiment is run five times. Then, we report the average scores with standard deviations. For all experiments, we select the best set of hyper-parameters from below:
\begin{itemize}
    \item Learning rate: \{$2\times 10^{-5}$, $3\times 10^{-5}$\};
    \item Number of epochs: \{2, 3, 4, 6, 8\};
\end{itemize}

\subsection{Dataset Statistics}

We truncate the instances of original Trivia dataset~\citep{joshi-etal-2017-triviaqa} to different lengths, to investigate whether Random Padding can improve QA models on datasets with various distributions of context length. 
\begin{table}[h]
\tabcolsep=0.3em
\centering
\begin{small}
\begin{tabular}{clccc}
\toprule
\multirow{2}*{Dataset} & \multirow{2}*{Split} & Avg. Length & Avg. Length & Num. of\\
& & of Questions & of Contexts & Instances\\
\midrule
\multirow{3}*{$\mathcal{D}_{800}$} & Train & 17.47 & 746.32 & 110,647\\
& Val & 17.18 & 732.97 & 4,924\\
& Test & 17.24 & 732.58 & 4,925\\
\midrule
\multirow{3}*{$\mathcal{D}_{100\sim 800}$} & Train & 17.47 & 275.17 & 110,647\\
& Val & 17.20 & 316.58 & 4,706\\
& Test & 17.14 & 317.14 & 4,709\\
\midrule
$\mathcal{D}_{100}$ & Train & 17.47 & 98.03 & 110,647\\
\midrule
$\mathcal{D}_{250}$ & Train & 17.47 & 242.34 & 110,647\\
\midrule
$\mathcal{D}_{400}$ & Train & 17.47 & 384.07 & 110,647\\
\bottomrule
\end{tabular}
\caption{\label{apptab:data_stat}
Statistics of all datasets adopted in Section \ref{sec:results_4_3}.
}
\end{small}
\vskip -0.15in
\end{table}

Since we mainly focus on whether QA models can find the precise boundary of answer span from the whole context, we remove the unanswerable instances from validation and test sets. However, to guarantee all training sets have the same scale, we do not remove any instances from training sets. We employ the function \texttt{word\_tokenize} of NLTK~\citep{nltk} to split questions and contexts into words. Then, we can confirm how many words a question or a context contains. Dataset statistics are provided in Table \ref{apptab:data_stat}. 

\section{Segment Division for the Analysis of Answer Position}
\label{app:seg_division}

To investigate the relationship between improvement from Random Padding and answer positions, we divide test set into two segments. The first segment consists of instances whose answers can be found at the front positions of contexts. And accordingly, the second segment consists of the instances whose answer are all located at the end of contexts.

We notice that there can exist more than one answer annotation in a context, and if models can find any mention of the answer, it will be considered as correct prediction. 
Thus, our division of dataset $\mathcal{S}$ is detailed as below:
\begin{itemize}
    \item We first assign a demarcation of answer positions, regarded as $X$.
    \item For every instance of dataset $\mathcal{S}$, if there exists at least one answer mention located at the position between $0$ and $X$, this instance will belong to the \textit{First Segment}, which is denoted as $\mathcal{S}^{(1)}_{X}$.
    \item The \textit{Second Segment} contains the rest instances of $\mathcal{S}$. Among these instances, all answer mentions are located between $X+1$ and $512$. Similarly, we denote this segment as $\mathcal{S}_{X}^{(2)}$.
\end{itemize}

For example, we divide a test set by the 100-th position. Supposing the answer of an instance is mentioned two times in context, located at 99 and 200, this instance should belong to the \textit{First Segment}. It is because there exist at least an answer annotation located between 0 and 100. However, if the two annotations are located at 101 and 200, this instance will belong to the \textit{Second Segment}, since models have to find answer from the rear tokens.

\section{Low-Resource Scenario}

We argue that Random Padding can make the rear part of position embeddings to be updated more times. However, if a model is trained on fixed dataset and fixed epochs, the total updating times of all position embeddings will be determinate, which means the updating times of front position embeddings will decrease after employing Random Padding. We wonder whether Random Padding can still be effective in low-resource scenario, where all position embeddings may not be properly trained. In this scenario, Random Radding might deteriorate model performance when predicting answers at the front positions, which might further hurt overall performance on the whole dataset. 

We randomly select sub-sets of $\mathcal{D}_{100}^{train}$ to train a QA model based on BERT-base. We create different degrees of low-resource scenarios, in which the training sets contain 5\%, 10\%, 25\%, 50\% instances of original training set, respectively. Similar to experiments mentioned above, we employ Adam~\citep{adam} as optimizer with learning rate of 3e-5. We sample instances for low-resource dataset with three different random seeds, and then experiment on them respectively. 

Similar to Section \ref{sec:inv_model}, we suppose there exist $M_q$ tokens in \textit{question} and $M_c$ tokens in \textit{context} after pre-processing. There also exist a \texttt{[CLS]} token and two \texttt{[SEP]} tokens in the input sequence. If the pre-trained language model recieve an input sequence with $n$ tokens, we should add $n-M_q-M_c-3$ padding tokens behind the \textit{context}. When training a model enhanced by Random Padding, we randomly select $k\in\left[0,\ n-M_q-M_c-3\right]$ and move $k$ padding tokens to the position between \texttt{[CLS]} and \textit{question}. In the low-resource scenario, we believe the number of moved padding tokens should be limited to $K$, to guarantee the model performance when answers are locating at the front position. Then, Random Padding can be modified as: When training, we select $k\in\left[0,\ \min\left\{K, n-M_q-M_c-3\right\}\right]$, and then move $k$ padding tokens to the front of \textit{question} tokens.

We empirically set $K$ to $64$, $128$, $192$, or $256$. Table~\ref{apptab:low_res} shows the results of models trained on low-resource dataset sampled with different random seeds. It shows that when the scale of training set become smaller, we should properly decrease the maximum number of moved padding tokens. For example, for the datasets sampled with Seed=42, the Random Padding model with $K=64$ achieves the best performance when model is trained with 5\% instances, while model with $K=128$ achieves the second best performance.  

\begin{table*}[t]
\centering
\tabcolsep=0.3em
\begin{small}
\subfigure[Sampling with Seed=42]{
\begin{tabular}{ccccc}
\toprule
Scale & $K$ & Eval F1 & Test F1 & $\Delta$Test F1\\
\midrule
\multirow{6}*{5\%} & $0$ & 47.49$_{\text{1.23}}$ & 47.17$_{\text{1.26}}$ & --\\
& $64$ & 48.72$_{\text{0.86}}$ & 48.94$_{\text{0.86}}$ & \tba{+1.77} \\
& $128$ & 49.16$_{\text{0.69}}$ & 48.86$_{\text{0.60}}$ & \tbb{+1.69} \\
& $192$ & 48.82$_{\text{0.84}}$ & 48.65$_{\text{0.71}}$ & \tbd{+1.48} \\
& $256$ & 48.89$_{\text{0.72}}$ & 48.84$_{\text{0.81}}$ & \tbc{+1.67} \\
& $512$ & 48.89$_{\text{0.76}}$ & 48.62$_{\text{0.84}}$ & \tbe{+1.45} \\
\midrule
\multirow{6}*{10\%} & $0$ & 52.50$_{\text{0.33}}$ & 52.29$_{\text{0.65}}$ & --\\
& $64$ & 53.04$_{\text{0.48}}$ & 52.90$_{\text{0.44}}$ & \tbc{+0.61}\\
& $128$ & 53.19$_{\text{0.16}}$ & 52.72$_{\text{0.34}}$ & \tbe{+0.43}\\
& $192$ & 53.03$_{\text{0.19}}$ & 52.97$_{\text{0.54}}$ & \tbb{+0.68}\\
& $256$ & 53.22$_{\text{0.33}}$ & 52.81$_{\text{0.69}}$ & \tbd{+0.52}\\
& $512$ & 53.10$_{\text{0.37}}$ & 53.28$_{\text{0.54}}$ & \tba{+0.99}\\
\midrule
\multirow{6}*{25\%} & $0$ & 56.30$_{\text{0.36}}$ & 56.39$_{\text{0.65}}$ & --\\
& $64$ & 56.82$_{\text{0.16}}$ & 56.89$_{\text{0.41}}$ & \tbe{+0.50}\\
& $128$ & 56.96$_{\text{0.33}}$ & 57.09$_{\text{0.31}}$ & \tbc{+0.70}\\
& $192$ & 56.89$_{\text{0.26}}$ & 57.00$_{\text{0.33}}$ & \tbd{+0.61}\\
& $256$ & 56.84$_{\text{0.13}}$ & 57.16$_{\text{0.52}}$ & \tbb{+0.77}\\
& $512$ & 57.06$_{\text{0.45}}$ & 57.30$_{\text{0.29}}$ & \tba{+0.91}\\
\midrule
\multirow{6}*{50\%} & $0$ & 59.37$_{\text{0.16}}$ & 59.89$_{\text{0.41}}$ & --\\
& $64$ & 60.09$_{\text{0.09}}$ & 60.24$_{\text{0.52}}$ & \tbe{+0.35}\\
& $128$ & 60.22$_{\text{0.19}}$ & 60.45$_{\text{0.26}}$ & \tbc{+0.56}\\
& $192$ & 60.26$_{\text{0.30}}$ & 60.29$_{\text{0.40}}$ & \tbd{+0.40}\\
& $256$ & 60.32$_{\text{0.14}}$ & 60.61$_{\text{0.35}}$ & \tba{+0.72}\\
& $512$ & 60.21$_{\text{0.17}}$ & 60.57$_{\text{0.51}}$ & \tbb{+0.68}\\
\bottomrule
\end{tabular}
}
\subfigure[Sampling with Seed=43]{
\begin{tabular}{cccc}
\toprule
$K$ & Eval F1 & Test F1 & $\Delta$Test F1\\
\midrule
$0$ & 47.18$_{\text{0.96}}$ & 46.82$_{\text{1.03}}$ & --\\
$64$ & 48.26$_{\text{0.78}}$ & 48.04$_{\text{0.95}}$ & \tbe{+1.22}\\
$128$ & 48.63$_{\text{0.61}}$ & 48.17$_{\text{0.95}}$ & \tbd{+1.35}\\
$192$ & 48.58$_{\text{0.48}}$ & 48.62$_{\text{1.08}}$ & \tbc{+1.80}\\
$256$ & 48.70$_{\text{0.66}}$ & 48.71$_{\text{0.61}}$ & \tbb{+1.89}\\
$512$ & 48.77$_{\text{0.59}}$ & 48.71$_{\text{0.57}}$ & \tba{+1.89}\\
\midrule
$0$ & 52.27$_{\text{0.74}}$ & 52.23$_{\text{0.98}}$ & --\\
$64$ & 52.93$_{\text{0.50}}$ & 52.89$_{\text{0.48}}$ & \tbd{+0.66}\\
$128$ & 52.99$_{\text{0.19}}$ & 52.91$_{\text{0.41}}$ & \tbc{+0.68}\\
$192$ & 52.74$_{\text{0.46}}$ & 52.46$_{\text{0.32}}$ & \tbe{+0.23}\\
$256$ & 53.16$_{\text{0.27}}$ & 53.16$_{\text{0.39}}$ & \tba{+0.93}\\
$512$ & 53.32$_{\text{0.49}}$ & 53.10$_{\text{0.34}}$ & \tbb{+0.87}\\
\midrule
$0$ & 56.38$_{\text{0.30}}$ & 56.44$_{\text{0.21}}$ & --\\
$64$ & 56.65$_{\text{0.36}}$ & 56.85$_{\text{0.57}}$ & \tbd{+0.41}\\
$128$ & 56.95$_{\text{0.22}}$ & 56.83$_{\text{0.32}}$ & \tbe{+0.39}\\
$192$ & 56.86$_{\text{0.39}}$ & 56.94$_{\text{0.40}}$ & \tbb{+0.50}\\
$256$ & 56.90$_{\text{0.30}}$ & 56.94$_{\text{0.28}}$ & \tba{+0.50}\\
$512$ & 56.96$_{\text{0.17}}$ & 56.92$_{\text{0.27}}$ & \tbc{+0.48}\\
\midrule
$0$ & 59.11$_{\text{0.28}}$ & 59.41$_{\text{0.09}}$ & --\\
$64$ & 59.47$_{\text{0.23}}$ & 59.81$_{\text{0.17}}$ & \tbe{+0.40}\\
$128$ & 60.05$_{\text{0.26}}$ & 60.20$_{\text{0.32}}$ & \tba{+0.79}\\
$192$ & 59.79$_{\text{0.09}}$ & 60.13$_{\text{0.61}}$ & \tbb{+0.72}\\
$256$ & 59.88$_{\text{0.28}}$ & 59.86$_{\text{0.30}}$ & \tbd{+0.45}\\
$512$ & 59.81$_{\text{0.18}}$ & 59.94$_{\text{0.54}}$ & \tbc{+0.53}\\
\bottomrule
\end{tabular}
}
\subfigure[Sampling with Seed=44]{
\begin{tabular}{cccc}
\toprule
$K$ & Eval F1 & Test F1 & $\Delta$Test F1\\
\midrule
$0$ & 47.20$_{\text{0.63}}$ & 46.63$_{\text{0.46}}$ & --\\
$64$ & 48.57$_{\text{0.69}}$ & 47.86$_{\text{0.60}}$ & \tbc{+1.23}\\
$128$ & 48.56$_{\text{0.61}}$ & 48.37$_{\text{0.74}}$ & \tba{+2.03}\\
$192$ & 48.64$_{\text{0.60}}$ & 47.69$_{\text{0.87}}$ & \tbe{+1.06}\\
$256$ & 48.43$_{\text{0.58}}$ & 47.83$_{\text{0.65}}$ & \tbd{+1.20}\\
$512$ & 48.67$_{\text{0.68}}$ & 48.01$_{\text{0.75}}$ & \tbb{+1.38}\\
\midrule
$0$ & 51.92$_{\text{0.52}}$ & 51.93$_{\text{0.74}}$ & --\\
$64$ & 52.46$_{\text{0.26}}$ & 52.13$_{\text{0.55}}$ & \tbe{+0.20}\\
$128$ & 52.49$_{\text{0.37}}$ & 52.36$_{\text{0.63}}$ & \tbd{+0.43}\\
$192$ & 52.59$_{\text{0.27}}$ & 52.58$_{\text{0.24}}$ & \tbb{+0.65}\\
$256$ & 52.60$_{\text{0.62}}$ & 52.83$_{\text{0.60}}$ & \tba{+0.90}\\
$512$ & 52.61$_{\text{0.26}}$ & 52.53$_{\text{0.34}}$ & \tbc{+0.60}\\
\midrule
$0$ & 56.63$_{\text{0.38}}$ & 56.68$_{\text{0.42}}$ & --\\
$64$ & 56.94$_{\text{0.25}}$ & 56.94$_{\text{0.49}}$ & \tbc{+0.26}\\
$128$ & 57.17$_{\text{0.38}}$ & 56.99$_{\text{0.26}}$ & \tbb{+0.37}\\
$192$ & 56.94$_{\text{0.52}}$ & 56.82$_{\text{0.29}}$ & \tbd{+0.14}\\
$256$ & 57.17$_{\text{0.23}}$ & 56.79$_{\text{0.23}}$ & \tbe{+0.11}\\
$512$ & 57.11$_{\text{0.45}}$ & 57.17$_{\text{0.43}}$ & \tba{+0.49}\\
\midrule
$0$ & 59.25$_{\text{0.43}}$ & 59.52$_{\text{0.52}}$ & --\\
$64$ & 59.69$_{\text{0.16}}$ & 59.78$_{\text{0.23}}$ & \tbd{+0.26}\\
$128$ & 59.71$_{\text{0.40}}$ & 59.91$_{\text{0.26}}$ & \tbc{+0.39}\\
$192$ & 59.81$_{\text{0.34}}$ & 59.74$_{\text{0.35}}$ & \tbe{+0.22}\\
$256$ & 59.96$_{\text{0.27}}$ & 60.05$_{\text{0.31}}$ & \tbb{+0.53}\\
$512$ & 59.91$_{\text{0.27}}$ & 60.07$_{\text{0.25}}$ & \tba{+0.55}\\
\bottomrule
\end{tabular}
}
\end{small}
\caption{\label{apptab:low_res}
Model performance on the low-resource datasets with different scales. $K=0$ represents the performance of baseline models without Random Padding, and $K=512$ represents the performance of models with original Random Padding.
}
\vskip -0.15in
\end{table*}

\end{document}